\documentclass{article}

\usepackage{graphics, graphicx} 
\usepackage{amsmath} 

\usepackage{xcolor}
\usepackage{soul}

\usepackage{subfigure}
\usepackage{float}
\usepackage{hyperref}
\usepackage{multirow}
\usepackage{amsmath,amssymb}

\usepackage{authblk}

\usepackage[utf8]{inputenc}
\usepackage[english]{babel}

\usepackage[nottoc]{tocbibind}

\begin{document}

\title{Quantitative analysis of robot gesticulation behavior\thanks{*This work has been partially supported by the Basque Government (IT900-16 and Elkartek 2018/00114), the Spanish Ministry of Economy and Competitiveness (RTI 2018-093337-B-100, MINECO/FEDER,EU).}
}

\author[1]{Unai Zabala}
\author[1]{Igor Rodriguez}
\author[1]{Jos\'e Mar\'{\i}a Mart\'{\i}nez-Otzeta}
\author[1]{Itziar Irigoien}
\author[1]{Elena Lazkano}
\affil[1]{Department of Computer Science and Artificial Intelligence, Faculty of Informatics, University of Basque Country (UPV/EHU), 20018 Donostia {\tt\small (igor.rodriguez@ehu.eus)}}

\maketitle

\begin{abstract}
Social robot capabilities, such as talking gestures, are best produced using data driven approaches to avoid being repetitive and to show trustworthiness. However, there is a lack of robust quantitative methods that allow to compare such methods beyond visual evaluation. In this paper a quantitative analysis is performed that compares two Generative Adversarial Networks based gesture generation approaches. The aim is to measure characteristics such as fidelity to the original training data, but at the same time keep track of the degree of originality of the produced gestures. Principal Coordinate Analysis and procrustes statistics are performed and a new Fréchet Gesture Distance is proposed by adapting the Fréchet Inception Distance to gestures. These three techniques are taken together to asses the fidelity/originality of the generated gestures.


\end{abstract}

\section{Introduction}
\label{sec:intro}

Advances in social robots are widespread in robotic conferences and newspapers. Robots for entertainment and care need to show socially acceptable behavior and, at the same time, must act in a non repetitive/boring manner and show trustworthiness.
An effective social interaction between humans and robots requires these robots follow the social rules and expectations of human users. That effectiveness largely depends on the non-verbal capabilities the robot is able to show during such interaction and it can be crucial for the social connection with humans as it allows a more intuitive communication \cite{cerrato2017engagement}.  
Gestures (head, arms and hands movements) are used both to reinforce the meaning of the words and to express feelings through non-verbal signs.
Gestures can be produced using a variety of approaches. They (a) can be ``manually produced'' by manually editing trajectories, (b) can be learned by demonstration or (c) data driven generative approaches can be used \cite{beck2017body}. 
However, there is a lack of a standardized method and quantitative metric for evaluating that social feature of a robot.  
How do we evaluate the generated gestures? How do we compare different approaches? Often authors make use of questionnaires that may help to validate the acceptability of a robot behavior out of the laboratory. Indeed, we think that visual validation is the first tool every robot behavior developer uses, but it is far from being a neutral quantitative method. 
Besides, motion analysis can help to detect jerky movements. But none of these give us two properties that are desirable specially when using generative models for gesture generation: the fidelity with respect to the original data used for acquiring the model and the degree of novelty/originality the obtained model offers.

In this paper we want to present a quantitative analysis of generated gestures according to several measures. It is not our goal to define a thorough methodology, but to give the researcher some tools, based in tested approaches, to help her with the always difficult and presumably impossible task of assessing the quality of generated gestures in an objective manner.
In the experiment that is described in this paper we use Generative Adversarial Networks (GAN) to generate talking beat gestures from a set of captured samples for the Softbank's robot Pepper\footnote{https://www.softbankrobotics.com/us/pepper}, where two variables have to be taken into account: the capture method (MoCap) and the length of the unit of movements (UM, a parameter intrinsic to our system that will be defined later). The goal is to test if the generated gestures are similar to the original ones, but at the same time possess some degree of originality. As it can be inferred, these two goals are contradictory, so a trade-off is needed.
To measure the fidelity of the generated samples to the original ones we performed a Principal Coordinate Analysis (PCoA) over the original and generated samples for the two types of MoCap and different length of units of movements. To measure the originality, we calculated procrustes statistics.
Finally, we have defined a Fréchet Gesture Distance (FGD) which is inspired in the Fréchet Inception Distance (FID). Assuming that the balance between fidelity and originality comes with the smaller FGD measure, this allows us to select the most appropriate value for the parameter being analysed.

Thus, the contribution of the paper is as follows:
\begin{itemize}
    \item Principal Coordinate Analysis (PCoA): a statistical tool for exploring the structure of high dimensional data. We propose this analysis to measure the degree of fidelity with respect to the training data. 
    \item Procrustes statistics is applied to ensure that the model is able to offer some originality to the gestures generated. The adequateness of the new movements is corroborated by motion measures such as jerk and path lenghts. 
    \item A new Fréchet Gesture Distance (FGD) is defined by adapting the Fréchet Inception Distance (FID) to the problem of GAN generated gestures. 
\end{itemize}

The rest of the paper is structured as follows: 
Section~\ref{sec:evaluation-overview} introduces the need for robot gesticulation and summarizes the different social skills evaluation alternatives found in the literature. Section~\ref{sec:experimental-baseline} describes the experimental baseline, the two GAN based gesture generation approaches that will be quantitatively analysed later on. The fidelity analysis is performed in Section~\ref{sec:similarity-analysis} while the originality analysis is described in Section~\ref{sec:originality-analysis}. The definition of the FGD measure is introduced in Section~\ref{sec:tradeoff} and this section also shows how the trade off has been conducted by calculating the distance between the generated gestures and the Gaussian Mixture Model (GMM) generated from a set of synthetic gestures created using Choregraphe, a software that allows to create robot animations. A qualitative visual evaluation is provided in Section~\ref{sec:visual-evaluation}.
Finally, Section~\ref{sec:further-work} is dedicated to the conclusions and to outline further work.

\section{Robot gesticulation. Evaluation alternatives}
\label{sec:evaluation-overview}

Talking  involves  spontaneous  gesticulation;  postures  and movements  are  relevant  for  social  interactions  even  if  they are  subjective  and  culture  dependent.  As co-thought (movements related to thinking activity) supports complex problem solving, co-speech implies communication~\cite{eielts2020cothought}. 
Lhommet and Marsella~\cite{lhommet15expressing} discuss body expression in terms of postures, movements and gestures. Gestures, defined as movements that convey information intentionally or not, are categorised as emblems, illustrators and adaptors. Emblems are gestures deliberately performed by the speaker that convey meaning by themselves and are again culture dependent. Illustrators are gestures accompanying speech, that may (emblems, deictic, iconic and metaphoric) or may not (beats) be related to the semantics of the speech \cite{mcneill1992hand}. Lastly, adaptors or manipulators belong to the gesture class that does not aid in understanding what is being said, such as ticks or restless movements.
Aiming  at  building trust and making people feel confident when interacting with them, socially acting humanoid robots should show human-like talking gesticulation. 

Problems arise when it comes to evaluate the behavior or a particular skill, e.g., the gesticulation ability, of a social robot. Usually robot behaviour is qualitatively evaluated. 
Often questionnaires are defined so that participants can rank several aspects of the robot's performance. There seems to be a consensus in presenting the questions using Likert scale and analyzing the obtained responses using some statistical test like analysis of variance, chi-square and so on.
For instance, in~\cite{velner2020intonation} social engagement with a robot can be evaluated by observing expressed emotions during the conversation. Humans participate in conversations with a NAO robot in different  intonation conditions. As objective measures they use number of turns  between actors, number of re-prompts, number of interruptions and the average silence length between turns. These measures are complemented with other subjective data such as the conversational naturalness, measured using questionnaires.
In~\cite{suguitan2020moveae} authors propose a method for modifying affective robot movements using neural networks. Again, the approach is evaluated using an online survey and Two one-sided tests (TOST).
Wolfert et al.~\cite{kucherenko2019importance} replicate the evaluation approach in~\cite{hasegawa2018evaluation} and assess the naturalness, semantic consistency and time consistency of the gestures generated by a speech driven encoder-decoder DNN performing a user-study. Once more, Becker-Asano and Ishiguro~\cite{becker2011evaluating} use questionnaires to investigate if facial displays of emotions with Geminoid F can be recognized and to find intercultural differences in the perception of those facial displays. Confusion matrices of the recognition rates are shown as measure.

Carpinella et al.~\cite{carpinella2017rosas} go one step further by developing a 18-item scale (based on psychological literature on social perception) to measure people's judgment of the social attributes of robots. This scale is also used in~\cite{pan2018evaluating} to examine how human collaborators perceive their robotic counterparts from a social perspective during object handovers.

When it comes to compare different approaches, data driven approaches are confronted to the original data that was used to learn the model and ranked results are then compared using some statistical tests. For instance, in~\cite{wolfert2019should} generated beat gestures are compared with designed beat gestures, timed beat gestures and noisy gestures using such approach. 

Qualitative methods are essential but are difficult to perform because a large number of evaluators is required and their subjective perceptions might be different. Moreover, when a large number of gestures must be evaluated the human eye becomes used to what is observing and it gets hard to remark the differences.
Thus, such methods are prone to result in subjective evaluation.  Besides, the evaluation is cultural dependent.

On the contrary, quantitative methods can handle a huge number of data as input, what makes them more appropriate to evaluate the robustness of a feature. However, subtle and subjective properties might not be easily measured with numerical methods. They cannot answer questions like "which one do you like it more?" neither can take into account the impact or effect a gesture system might have on a specific target audience.  
Both evaluation methods have strengths and weaknesses and are complementary.

Rare are the references that use quantitative evaluation methods.
In~\cite{rodriguez2019spontaneous} gestures generated by a GAN network are compared with gestures obtained by GMM, Hidden Markov Model (HMM) and gestures obtained by randomly ordering the training data. Principal coordinates analysis was used to extract the similarities between the generated gestures and the original ones. Other features such as 3D space coverage, path length and motion jerk were also used for evaluation purposes. Similar motion statistics were used in~\cite{kucherenko2020gesticulator}. More specifically, the average values of the root-mean-square error and speed histograms of the produced motion are shown as new measures.

Social behavior must be socially acceptable  above all and questionnaires are very valuable tools that need to be considered. But when it comes to compare several approaches objective tools are needed. We have focused on three characteristics (from the seven ones stated in \cite{borji2019pros}) as desirable when using a data-driven gesture generation approach:

\begin{itemize}
    \item Ability to generate high fidelity samples
    \item Ability to generate diverse samples
    \item Agreement with human perceptual judgments and human rankings of models
\end{itemize}

These characteristics could be in contradiction among them, particularly the fidelity and diversity constraints. We have tried our best to try to reconcile them and after the quantitative analysis we have returned the human to the loop for the test of the third condition: human judgement.
Therefore, in the research described in this paper we perform an analysis based on several methods for quantitatively measuring the degree of fidelity as well as originality a gesture generation method offers with respect to the properties of the data used for training the system.

\section{Experimental baseline}
\label{sec:experimental-baseline}
 
In this section the two GAN based gesture generation methods that will be quantitatively analysed later on are explained. Both methods can generate human-like motion in a humanoid robot that includes arms, head and hands motion mapping (upper-body part) since legs are not involved in talking beats, and only differ in the 3D MoCap system (OpenNI vs OpenPose\cite{cao2018openpose}) being used to capture human motion and create the databases for acquiring the generative models.
The motion capturing alternative over already existing robot animation software clearly allows to better capture the nature of the talking movements we do, but it requires the capability to (1) Capture good features of the motion (2) Map those captured features into the robot joints~\cite{penna2013whole}. 
 This mapping process can be done by inverse kinematics \cite{alibeigi2017inverse}, calculating the necessary joint positions given a desired end effector's pose as in~\cite{mukherjee2015inverse}. 
Alternatively, we adopt the direct kinematics option that straightforwardly adapts the captured arms and head angles to the robot joints \cite{zhang2019real}
\cite{rodriguez2014humanizing}. 

\begin{figure}[!htbp]
  \includegraphics[width=0.8\columnwidth]{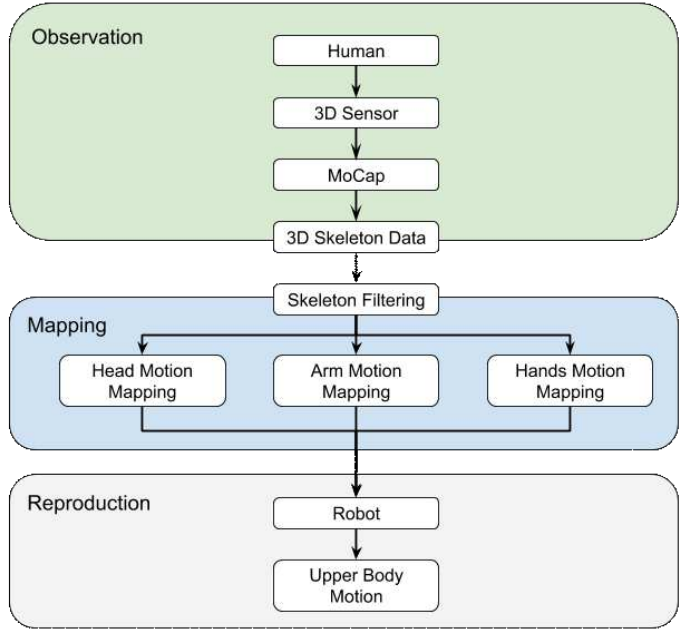}
  \caption{Human motion imitation process}
  \label{fig:imitation-process}
\end{figure}

The mapping process leads to the capability of human motion imitation as depicted in
Figure~\ref{fig:imitation-process}. 

\subsection{Mapping human joints to robot joints: OpenNI vs OpenPose}
\label{subsec:mapping-joints}
\begin{figure}[!htbp]
   \subfigure[OpenNI\label{fig:openni-skel}]{
     \includegraphics[width=0.25\columnwidth]{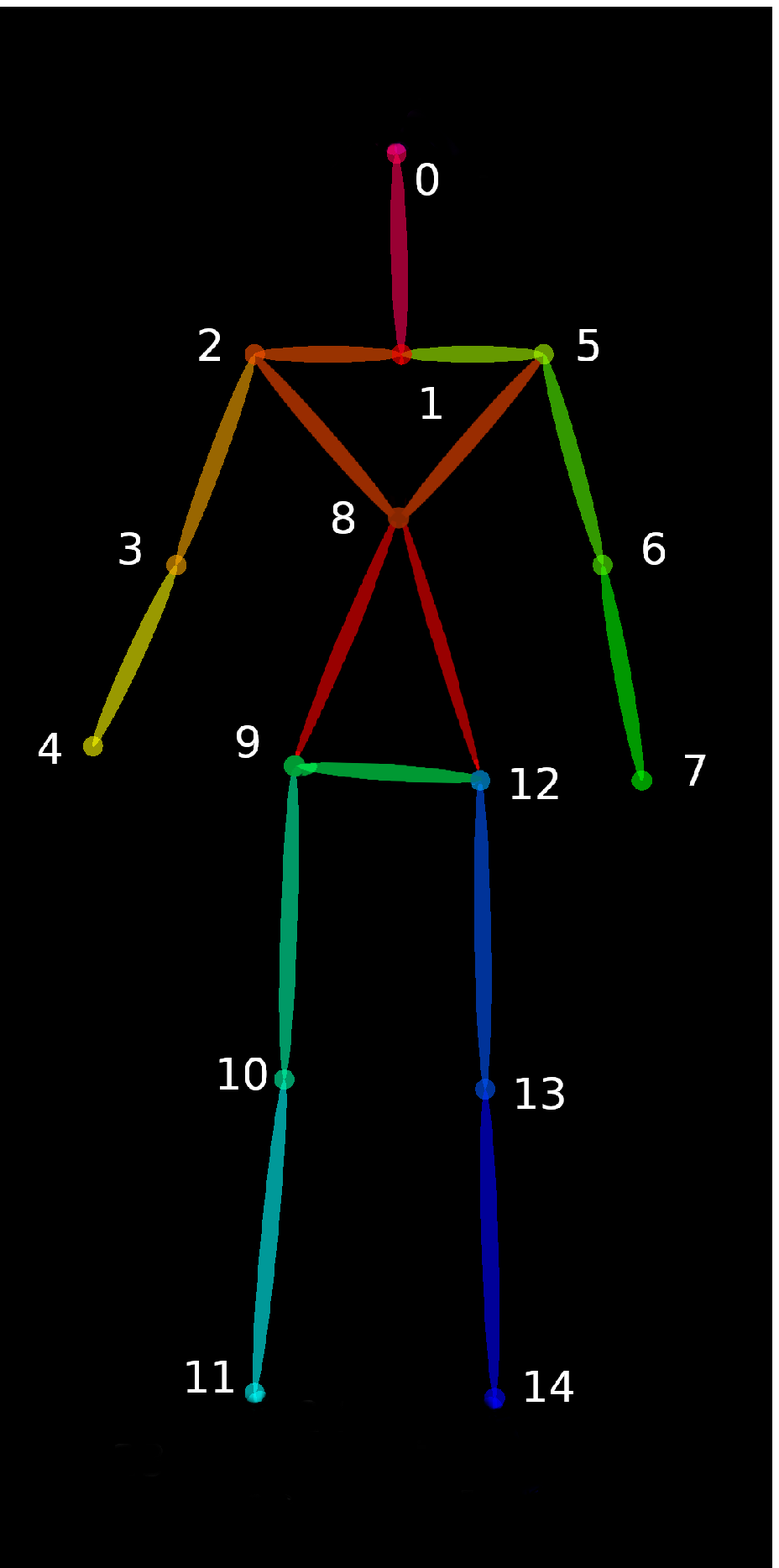}}
   \subfigure[OpenPose\label{fig:openpose-skel}]{
     \includegraphics[width=0.25\columnwidth]{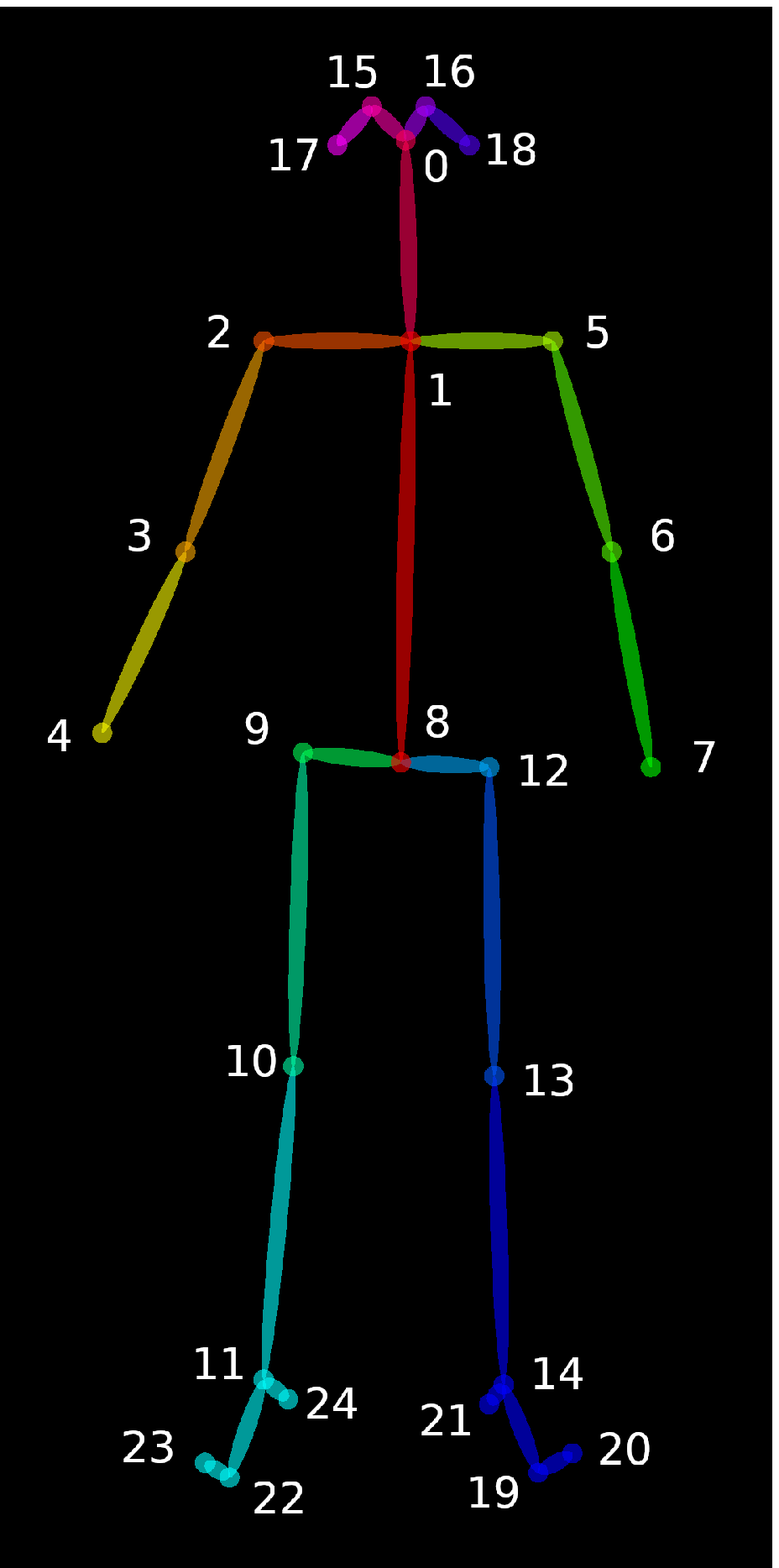}}
\caption{OpenNI and OpenPose skeleton models}
\label{fig:openni-openpose-skels}
\end{figure}

As mentioned before, arms, head and hands are involved in the gesture generation process. The MoCap systems being used show different features and limitations and thus, the mapping process differs from one to the other in some joints. More specifically, head and hands need to be differently mapped. Figure~\ref{fig:openni-openpose-skels} reflects the main differences between these two systems. OpenNI can only detect 15 keypoints while OpenPose detects 25 for the body plus the 42 hands' keypoints (hands are not displayed in the figure aforementioned). The following subsections detail how those elements are translated from human captured 3D cartesian coordinates to the joints of Softbank's robot Pepper.

\paragraph{Arms mapping}
The literature reveals different approaches to calculate the robot arm joint positions~\cite{zhang2019real}\cite{kofinas2015complete}. This mapping process depends upon the robot's degrees of freedom and joints range. For the Pepper robot arms we are dealing with, some upper-body link vectors are built through the skeleton points in the human skeleton model, and joint angles are afterwards extracted from the calculation of the angles between those vectors. For the sake of simplicity, since the calculation of the angles is similar for both approaches, the complete formulas involved in that process will not be described here (see~\cite{zabala2019learning} for more detailed information).

\begin{figure}[!htbp]
\centering
\includegraphics[width=1.0\columnwidth]{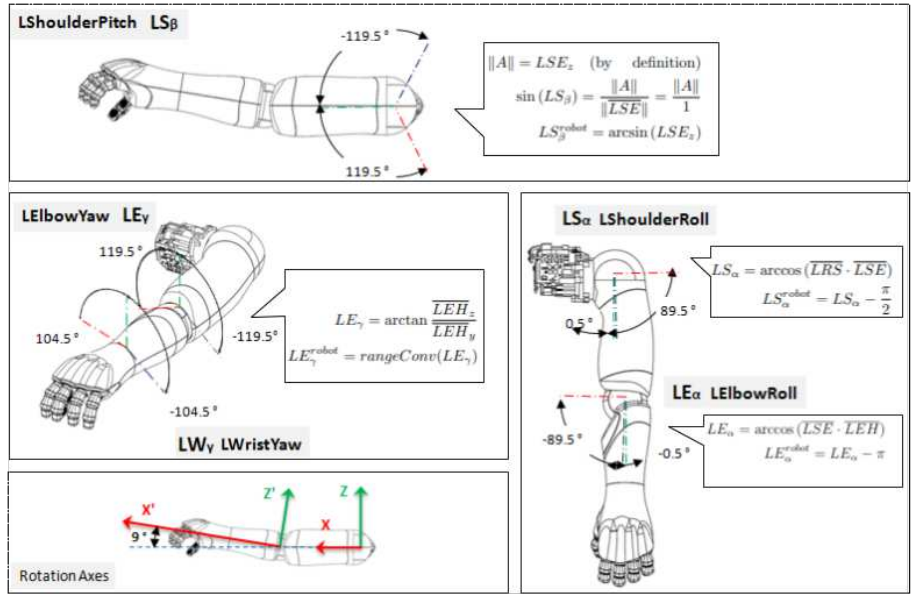}
\caption{Left arm joints and angle limits}
\label{fig:left-arm}
\end{figure}

\paragraph{Head mapping}
The OpenNI skeleton tracking program employed for head mapping gives us the neck and head 3D poses. The approach taken for mapping the yaw angle to the robot's head consists of applying a gain $K_1$ to the human's yaw value, once transformed into the robot space by a $-\frac{\pi}{2}$ rotation (Eq.~\ref{eq:openni_head_yaw}).

\begin{equation}
H^{robot}_\gamma = K_1 \times H_\beta 
\label{eq:openni_head_yaw}
\end{equation}

In order to approximate head's pitch angle, 
the head to neck vector ($\overline{HN}$) is calculated and rotated $-\frac{\pi}{2}$ and then,
its angle is obtained
(Eq.~\ref{eq:openni_head_pitch}). Note that robot's head is an ellipsoid
instead of an sphere. To avoid unwanted head movements a lineal gain
is applied to the final value.

\begin{equation}
\label{eq:openni_head_pitch}
H^{robot}_\beta = \arctan{(rotate(\overline{HN}, -\frac{\pi}{2}))} + |K_2*H_\gamma|
\end{equation}

On the contrary,
OpenPose detects basic face features such as the nose, the eyes and the ears (see Figure \ref{fig:openpose-replica}) and thus, allows for a more realistic tracking of the robot head. To map humans head position into the robot, we use the nose position as reference. Head's pitch $(H^{robot}_\phi)$ is proportional to the distance between the nose and the neck joint (see Eq.~\ref{eq:openpose_head_pitch}). Instead, the yaw orientation of the head itself $(H^{robot}_\psi)$ can be calculated by measuring the angle between the vector joining the nose and the neck, and the vertical axis (see Eq.~\ref{eq:openpose_head_yaw}).

\begin{equation}
    \begin{split}
    \overline{NN} = dist(Nose, Neck) \\
    H^{robot}_\phi = rangeConv(\overline{NN}, robotRange) \\
    \end{split}
    \label{eq:openpose_head_pitch}
\end{equation}

\begin{equation}
    \begin{split}
    H^{robot}_\psi = rangeConv(-\arcsin{(\overline{NN}_x)}, robotRange)
\\
    \end{split}
    \label{eq:openpose_head_yaw}
\end{equation}

\paragraph{Hands mapping}
The OpenNI skeleton tracking program used for hands mapping can not detect the operator's hands' yaw motion and thus, $LW_\gamma$ joints cannot be reproduced using the skeleton information.  
The developed solution forces the user to wear coloured gloves to detect the hand orientation. In our implementation, the gloves are green in the palm of the hand and red in the back. While the human talks, hands coordinates are tracked and those positions are mapped into the image space and a subimage is obtained for each hand. Angular information is afterwards calculated by measuring the number of pixels ($max$) of the outstanding color in a subimage. Eq.~\ref{eq:hands} shows the procedure for the left hand. $N$ is a normalizing constant and $maxW_\gamma$ stands for the maximum wrist yaw angle of the robot.

\begin{equation}
\left\{ 
\begin{array}{ll} LW^{robot}_\gamma =   max/N \times maxW_\gamma & \mbox{ if $max$ is palm}\\
LW^{robot}_\gamma =  \frac{max-N}{N} \times maxW_\gamma
  & \mbox{ otherwise}
  \end{array}\right.
\label{eq:hands}
\end{equation}

In addition, $LE_\gamma$ is modified when humans palms are up to ease the movement of the robot. 

Regarding the fingers, as they cannot be tracked, their position is randomly set at each skeleton frame to make the movement more realistic.

\begin{figure}[!htbp]
 \includegraphics[width=0.4\columnwidth]{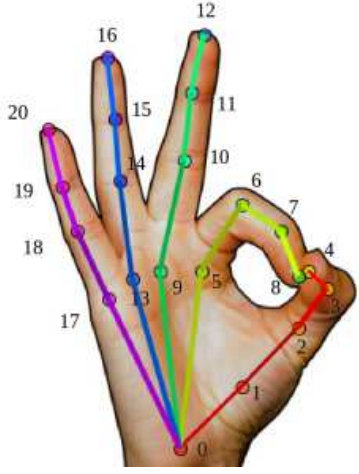}
 \caption{OpenPose hand model (21 keypoints)}
 \label{fig:hands-coco}
\end{figure}

Alternatively, OpenPose differentiates left and right sides without any calibration and gives 21 keypoints per hand, four per finger plus wrist (see Figure~\ref{fig:hands-coco}). 

    To determine if the hand shows the palm or the back the angle between the horizontal line and the line joining the thumb and the pinky fingertips (keypoints 4 and 20) is required. This calculation is expressed in Eq.~\ref{eq:hand-side-right}, where $FT$ stands for fingertip and $OFT$ represents the new origin of a fingertip. Afterwards, the fingers' points are rotated in such a way that the pinky lies at the right of the thumb and the number of fingers over the $Y=0$ line is calculated (Eq.~\ref{eq:hand-side-right2}). For the right hand, at least two fingers should lie over that line to consider the palm is being showed as shown if Eq.~\ref{eq:hand-side-right3} (the opposite condition for the left hand).
    
    \begin{equation}
    \label{eq:hand-side-right}
    \begin{split}
    \forall i FT^i \ \ \ OFT^i_{x,y} = FT^i_{x,y} - Thumb_{x,y} \\
    \alpha = \arctan(OFT^{pinky}_y, OFT^{pinky}_x) \\
    \end{split}
    \end{equation}
    
    \begin{equation}
    \label{eq:hand-side-right2}
    \begin{split}
    \forall i \ \  FT'^{i}_{x} = OFT^i_{x} * \cos{(-\alpha)} - OFT^i_{y} * \sin{(-\alpha)} \\
    \forall i \ \  FT'^{i}_{y} = OFT^i_{x} * \sin{(-\alpha)} - OFT^i_{y} * \cos{(-\alpha)} \\
    \end{split}
    \end{equation}
    
    \begin{equation}
    \label{eq:hand-side-right3}
    \begin{split}
     HandSide = \begin{cases}
       Back &  ((\sum_{i=1}^{3} FT'^{i}_y) > 0) \geq 2 \\
       Palm & otherwise \\
      \end{cases}
    \end{split}
    \end{equation}
    
    In addition, each hand's yaw angle $(H_\psi)$ must be calculated by measuring the distance between the thumb and the pinky fingertips (Eq.~\ref{eq:hand_yaw}). The minimum and maximum values are adjusted according to the wrist's height so to avoid collisions with the touch screen on the chest of the robot. 
    \begin{equation}
    \begin{split}
    \overline{TP} = dist( FT'^{Thumb}, FT'^{Pinky}) \\
    H_\psi = rangeConv(\overline{TP}, robotRange)
    \end{split}
        \label{eq:hand_yaw}
    \end{equation}
    
    Finally, the hand's opening/closing is defined as a function of the distance between wrist (keypoint 0) and middle fingertip (keypoint 12) as in Eq.~\ref{eq:hand_opening}. 
    \begin{equation}
    \begin{split}
    \overline{MW} = dist(FT^{Middle}, Wrist) \\
    ClosedOpen = rangeConv(\overline{MW}, [0.0-1.0])
    \end{split}
        \label{eq:hand_opening}
    \end{equation}

\color{black}
\subsection{GAN based gesture generation}
\label{subsec:gan-gesture-generation}
 GAN networks are composed by two different interconnected networks. The \emph{Generator} (\textit{G}) network generates possible candidates so that they are as similar as possible to the training set. The second network, known as the \emph{Discriminator} (\textit{D}), judges the output of the first network to discriminate whether its input data are ``real'', namely equal to the input data set, or if they are ``fake'', that is, generated to trick with false data.

 As we are interested in generating movements, i.e., a sequence of poses, the input to the learning process to any generative model has to take into account the temporal sequence of poses. The training dataset given to the \textit{D} network contains K unit of movements (UM), being each UM a sequence of $\mu$ consecutive poses, and each pose 14 float numbers corresponding to joint values ($J_i$) of head, arms, wrists (yaw angle) and hands (finger opening value).
 
 Table~\ref{tab:db-shape} describes more in detail the aspect of a single entry of the database for the case of $\mu = 4$. These samples were recorded by using two different MoCap systems and by registering 10 different person talking about 18 minutes overall. Therefore, two datasets have been obtained: OpenNI DB is built from a recording about nine minutes long and contains five people's skeleton data captured using OpenNI as MoCap system, while OpenPose DB is built from another recording also about nine minutes long that contains other different five people's skeleton data captured using OpenPose. After sampling with a frequency of 4 Hz, two datasets of a slightly different number of poses are created. The shorter one has 2018 poses, and the last poses of the longer one have been deleted to make their lengths match.

 \begin{table}[!htbp]
\hspace{-1cm}
\fbox{
\(
\scriptsize
J_1(t) \cdots J_{14}(t), J_1(t+\Delta t) \cdots J_{14}(t+\Delta t), J_1(t+2\Delta t) \cdots J_{14}(t+2\Delta t), J_1(t+3\Delta t) \cdots J_{14}(t+3\Delta t) 
\)
}
\caption{Characterization of a unit of movement for $\mu =4$ consecutive poses. $\Delta t$ depends of the data sampling frequency}
\label{tab:db-shape}
\end{table}

\color{black} 

The \textit{D} network is thus trained using that data to learn its distribution space; its input dimension is $\mu * 14$. On the other hand, the \textit{G} network is seeded through a random input with a uniform distribution in the range [$-1, 1$] and with a dimension of 100. The \textit{G} network intends to produce as output gestures that belong to the real data distribution and that the \textit{D} network would not be able to correctly pick out as generated. 

\section{Fidelity analysis}
\label{sec:similarity-analysis}

Dimension Reduction techniques are very widely used in very different areas, such as in genomics, image classification or in natural language processing tasks. The most well known is the Principal Component Analysis (PCA) \cite{hotelling1993analysis} and it can help to explore the structure of high dimensional data. It is a technique that displays the structure of complex data in a high dimensional space into a lower dimensional space without too much loss of information. In robotics, particularly when studying motions or movements, PCA has also extensively been applied. \cite{park15} used PCA to build motions within an imitation learning framework;  \cite{wood18} used PCA to increase the interpretability of upper limb's movements registered by a robotic technology for different tasks; and in \cite{jarque19} data acquired with a dataglove was summarized with PCA to extract the coordination patterns available for handgrasps.
Principal Coordinates Analysis \cite{gower1966some} (PCoA), also known as Classical Multidimensional Scaling,  is an extension of the  PCA and therefore it allows to explore and visualize similarities or dissimilarities of data.  Given $n$ units and distances $d_{ij}$ between each pair of units $i$ and $j$, all the distances are  gathered in a $n \times n$ distance matrix $\mathbf{D}$. The PCoA builds a new matrix $\mathbf{Y}$ containing the coordinates of the $n$ units in $l$ dimensions such that the Euclidean distance between the $i$-th and $j$-th units is equal to $d_{ij}$ for all $i$ and $j$. The columns of matrix $\mathbf{Y}$ are given basically by the eigenvectors of the inner product matrix $(\mathbf{I} - \mathbf{1}\cdot\mathbf{1}'/n) \Tilde{\mathbf{D}}(\mathbf{I} - \mathbf{1}\cdot\mathbf{1}'/n)$, where $\Tilde{\mathbf{D}}$ is the matrix with value $(d_{ij})^2$ in position $(i, j)$, $\mathbf{1} = (1, \ldots, 1)'$ and $\mathbf{I}$ is the identity matrix. The related eigenvalues show the variability decomposition in the original data. When the distance matrix $\mathbf{D}$ is the Euclidean distance built on the original features, PCoA and PCA give the same results. In summary, the columns of matrix $\mathbf{Y}$ along with the eigenvalues allow to analyse the internal structure of the original high dimensional data.

\subsection{Measuring similarity with PCoA}
\label{subsec:sim-with-pcoa}
\begin{figure}[!htbp]
\includegraphics[width=0.99\columnwidth]{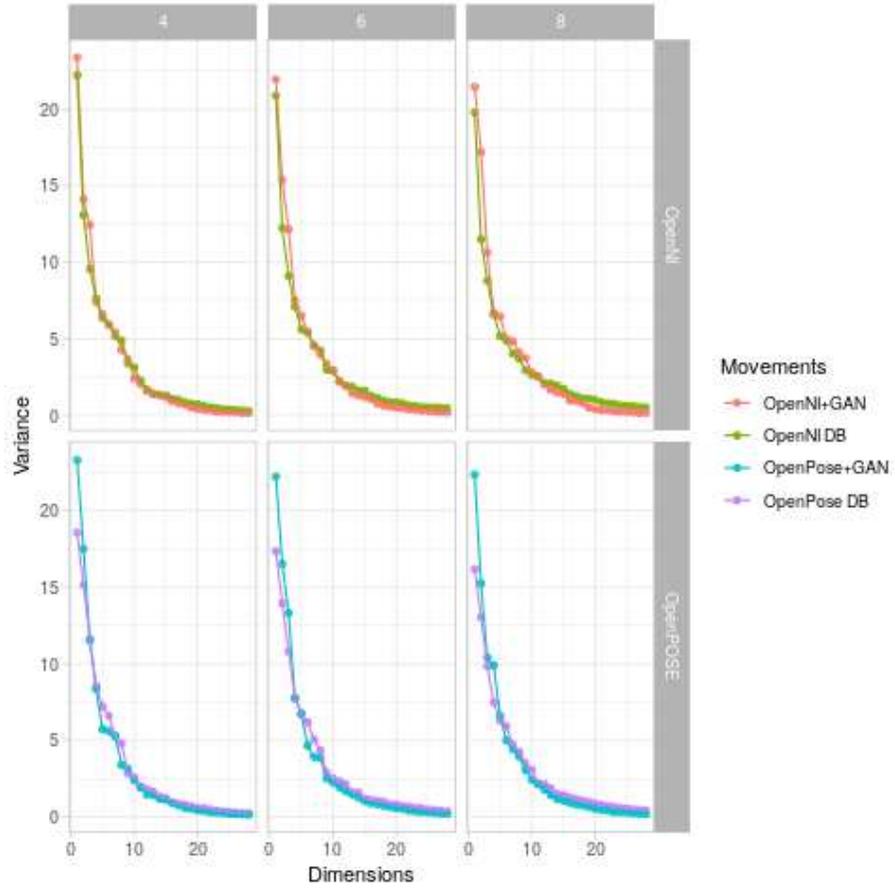}
\caption{Decomposition of the variance for different length of units and different systems of movement ($\lambda_{Ol}, \lambda_{Gl}, l=1 \ldots, 28$). The columns are ordered by length of UM ($\mu=4, 6, 8$). The first and second rows correspond with data acquisition by OpenNI and OpenPose, respectively.}
\label{fig:vp}
\end{figure}

\begin{figure}[!htbp]
\includegraphics[width=0.99\columnwidth]{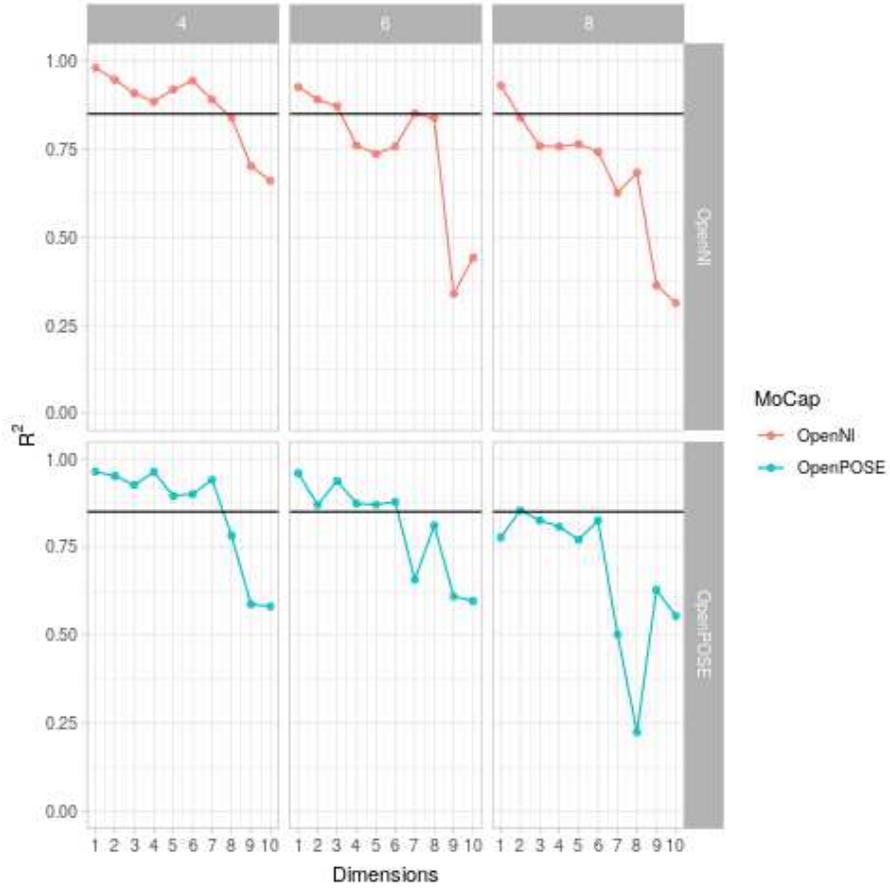}
\caption{Determination coefficients ($R^2$) for lineal models $\mathbf{y}_{jO} = \beta_0 + \sum_{i=1}^{10}\beta_i\mathbf{y}_{iG}$, $j=1, \ldots, 10$. The columns are ordered by length of UM ($\mu=4, 6, 8$). The first and second rows correspond with data acquisition by OpenNI and OpenPose, respectively.}
\label{fig:R2}
\end{figure}

Let OpenNI DB and OpenNI+GAN be the databases captured and generated respectively with the OpenNI capture method. The same holds for OpenPose DB and OpenPose+GAN. The databases
OpenNI DB, OpenNI+GAN, OpenPose DB, OpenPose+GAN were calculated for different length of UM ($\mu =4, 6, 8$). 
 This gives a $N  \, \times  \,(14\times \mu)$ data matrix for each method where columns represent the positions of the joints along the sequence  of $\mu$ consecutive  poses ($J_i(t + k\Delta t)\, , i=1, \ldots, 14, \; k=0, \ldots \mu-1, \; \mu=4, 6, 8$). The structure underlying the movements was analyzed considering the relationship between the joints. First, correlation distances~\cite{gower85encyclopedia} 
between joints  were calculated. In order to get comparable results the distance matrices were scaled so that their geometric variability were equal to 1. Then, a Principal Coordinates Analysis  was carried out on each distance matrix. In order to assess whether the underlying structures of original movements and generated movements are similar, the corresponding eigenvalues were compared.
Figure \ref{fig:vp} shows the decomposition of the variance (first 28 dimensions) for different lengths of UM and different methods of data acquisition. It can be seen that, in general terms, the structure of the original movements are preserved by the GAN generated samples, independently of the MoCap system being used. Furthermore, in order to assess quantitatively the fidelity between GAN generated samples and its originals, the principal coordinates $\mathbf{Y}_O$ of the originals and the principal coordinates $\mathbf{Y}_G$ of the GAN generated samples were compared. Particularly, we measured the ability to recover each of the first 10 principal coordinates $\mathbf{Y}_O = [\mathbf{y}_{1\,O}, \ldots, \mathbf{y}_{10\,O}]$ from the first 10 principal coordinates $\mathbf{Y}_G$ based on linear regression models. We considered the linear regression model for the $j$th principal coordinate of the originals based on the 10 principal coordinates of GAN generated samples ($\mathbf{y}_{jO} = \beta_0 + \sum_{i=1}^{10}\beta_i\mathbf{y}_{iG}$, $j=1, \ldots, 10$) and calculated the explained variance by the coefficient of determination $R^2$ (see Figure \ref{fig:R2}). Broadly, very high values of $R^2$ are obtained assessing the fidelity of GAN models in concordance with what the eigenvalue decompositions showed. Nevertheless, we gained some insight and it can be observed that the recovery of the originals is bigger with OpenPose as MoCap. Furthermore, the recovery for 8 UM is the poorer. It can be seen that first 6 and 7 principal coordinates of OpenPose DB can be recovered by the GAN principal coordinates ($R^2 \ge 0.85$) for $\mu$ 4 and 6, respectively.

\section{Originality analysis}
\label{sec:originality-analysis}
Nothing has been said about the degree of originality of the generated motion. As mentioned in the introduction, robot gesticulation should not result repetitive/boring. 

In order to analyze it, we considered Procrustes Analysis. Procrustes methods analyze the matching between two or more configurations. That is, given some units measured in different contexts or by different features, the main aim of procrustes methods \cite{gower2004procrustes} is to measure the degree of similarity among the configurations. Procrustes methods are widely applied. For instance, in \cite{makondo2015knowledge} the authors extend procrustes statistic to get transfer learning techniques to learn robot kinematic and dynamic models; in \cite{gao2016calibration}  applied procrustes techniques as an effective robot base frame calibration;  more recently, \cite{maset2020procrustes} proposed a method to increase efficiency and to identify potential issues of the assembly process in robotized assembly as a variation of the classical procrustes analysis.

\subsection{Measuring originality with procrustes statistic}
\label{sec:mesuring-originality}
In our particular context, we considered pairs of configurations given by
the first 10 principal coordinates $\mathbf{Y}_O$ of the originals and the 10 principal coordinates $\mathbf{Y}_G$ of the GAN generated samples for each combination of MoCap and UM. The rows of those matrices represent the joints along the unit of movement and the matrices can be considered as configurations for the joints. Based on the percentage of the explained variances (see Table \ref{tab:PCA_percentage}) the aforementioned configurations are capturing the essence of the joints along the units of movement. The classical orthogonal procrustes statistic ($ss$) between configurations $\mathbf{Y}_O$ (DB) and $\mathbf{Y}_G$ (GAN)  is the residual sum of squares between both configurations, once a scaling factor and rotation movement are allowed. That is, $ss= ||\mathbf{Y}_O - s Q \mathbf{Y}_G||^2$, where $s$ is a scaling factor and $Q$ is a rotation matrix that minimize the sum of squares. The underlying idea is to consider $\mathbf{Y}_O$ as the target configuration and to scale and rotate the second configuration $\mathbf{Y}_G$ so that it is as similar as possible to the target configuration. The remaining residuals build the procrustes statistic $ss$. The bigger is $ss$ the more different are the joints along the units of movements, or in our context, the bigger is the originality of the movements. Since $ss$ depends on the number of rows, we normalize it so that we obtain a commeasurable statistics for different length of UM. Taking into account Table \ref{tab:PCA_percentage}, in terms of originality it seems that MoCap OpenPose obtains slightly bigger values.

\begin{table}[!htbp]
\caption{Explained variance along the first 10 dimensions for different length of UM and different systems of movement. Differences between joints along UM measured by the commeasurable procrustes statistic $ss/(14 \mu)$.}
\label{tab:PCA_percentage}
\begin{tabular}{*{4}c}
\hline
 & \multicolumn{2}{c}{Explained variance (\%)} & \\ 
$\mu$ & Original & OpenNi & $ss/(14\times \mu)$\\
\hline
4 & 81.3 & 85.6 & 0.0857\\
6 & 75.0 & 83.7 & 0.1723\\
8 & 70.1 & 82.9 & 0.1932\\
\hline
$\mu$ & Original & OpenPose & \\
4 & 83.2 & 86.2 & 0.1054\\
6 & 77.6 & 83.8 & 0.1307\\
8 & 74.4 & 83.4 & 0.2369\\
\hline
\end{tabular}
\end{table}

The originality should not come at the cost of rough or uneven movements. Tables~\ref{tab:measures-sim-4} to~\ref{tab:measures-sim-8} show the mean values of the norm of jerk~\cite{calinon04learning} (equation~\ref{eq:jerk}) and the length of the path (described as the increment in the positions over time in equation~\ref{eq:lpath}) for 1000 generated movements. Head position does not shift in the space, and thus only jerk values are calculated.
Overall, motion analysis shows that OpenPose based gesture generation is smoother than the OpenNI based one, independent of the length of the unit of movement selected.

\begin{equation}
\label{eq:jerk}
{jerk} = \frac{1}{T}\sum_{t=1}^{T}||\dot{accel_t}||
\end{equation}

\begin{equation}
\label{eq:lpath}
{lpath} = \sum_{t=2}^{T}||{\overline{x}_t - \overline{x}_{t-1}}||
\end{equation}

  \begin{table}[h]
   \caption{Mean values for each measure ($\phi$: pitch, $\psi$: yaw) $\mu = 4$}
  \label{tab:measures-sim-4} 
\begin{tabular}[!htbp]{lc|cccc|}\hline
 & & OpenPose based & OpenNI based \\ \hline
 \multirow{2}{*}{Lhand} &  $E_{jerk} $ &   0.0266 & 0.0232\\
  &  $E_{lpath}$ & 25.8314 &  21.7854 \\  \hline
    \multirow{2}{*}{Rhand} &  $E_{jerk} $ & 0.0232 & 0.0231\\
  &  $E_{lpath}$ & 23.4980 & 21.2495 \\  \hline
 \multirow{2}{*}{Lelbow} &  $E_{jerk} $ &   0.0108 & 0.0118\\
  &  $E_{lpath}$ & 10.8245 &  11.9210 \\  \hline
    \multirow{2}{*}{Relbow} &  $E_{jerk} $ & 0.0086 & 0.0110\\
  &  $E_{lpath}$ &  9.1785 &  10.6696 \\  \hline
    \multirow{2}{*}{Head} &  $E^{\psi}_{jerk} $ & 0.0428 & 0.0334\\
  &  $E^{\phi}_{jerk}$  & 0.0201 &  0.0153 \\ \hline
  \end{tabular}
\end{table}

  \begin{table}[h]
    \caption{Mean values for each measure ($\phi$: pitch, $\psi$: yaw) $\mu = 6$}
  \label{tab:measures-sim-6}
\begin{tabular}[!htbp]{lc|cccc|}\hline
 & & OpenPose based & OpenNI based \\ \hline
 \multirow{2}{*}{Lhand} &  $E_{jerk} $ & 0.0240 & 0.0285\\
  &  $E_{lpath}$ & 22.4936 &  25.0769 \\  \hline
    \multirow{2}{*}{Rhand} &  $E_{jerk} $ & 0.0220 & 0.0273\\
  &  $E_{lpath}$ & 22.7787 & 26.0431 \\  \hline
 \multirow{2}{*}{Lelbow} &  $E_{jerk} $ & 0.0094 & 0.0132\\
  &  $E_{lpath}$ & 9.5669 &  11.6939 \\  \hline
    \multirow{2}{*}{Relbow} &  $E_{jerk} $ & 0.0083  & 0.0142\\
  &  $E_{lpath}$ &  9.1389 & 13.2678 \\  \hline
    \multirow{2}{*}{Head} &  $E^{\psi}_{jerk} $ & 0.0389 &0.0436\\
  &  $E^{\phi}_{jerk}$  & 0.0198 & 0.0131 \\ \hline
  \end{tabular}
\end{table}

  \begin{table}[h]
    \caption{Mean values for each measure ($\phi$: pitch, $\psi$: yaw) $\mu = 8$}
    \label{tab:measures-sim-8}  
\begin{tabular}[!htbp]{lc|cccc|}\hline
 & & OpenPose based & OpenNI based \\ \hline
 \multirow{2}{*}{Lhand} &  $E_{jerk} $ & 0.0244 & 0.0286\\
  &  $E_{lpath}$ & 24.2394 &  25.4619 \\  \hline
    \multirow{2}{*}{Rhand} &  $E_{jerk} $ & 0.0282 & 0.0374\\
  &  $E_{lpath}$ & 29.2930 & 33.4755 \\  \hline
 \multirow{2}{*}{Lelbow} &  $E_{jerk} $ & 0.0109 & 0.0145\\
  &  $E_{lpath}$ & 11.3923 &  13.7295 \\  \hline
    \multirow{2}{*}{Relbow} &  $E_{jerk} $ & 0.0131  & 0.0190\\
  &  $E_{lpath}$ &  13.7552 & 17.2916 \\  \hline
    \multirow{2}{*}{Head} &  $E^{\psi}_{jerk} $ & 0.0513 &0.0527\\
  &  $E^{\phi}_{jerk}$  & 0.0267 & 0.0122 \\ \hline
  \end{tabular}
\end{table}

\section{Trade off between fidelity and originality}
\label{sec:tradeoff}
As mentioned in the introduction, the fidelity and the originality features are contradictory and a trade-off is desirable. Looking for that balance, we have defined a Fréchet Gesture Distance, inspired by the Fréchet Inception Distance, a distance widely used in the area of image generation to measure the similarity between original images and GAN generated images.

\subsection{GAN performance metrics}
\label{subsec:gan-performance-metrics}

Evaluation of the performance of GAN networks is not a straightforward process. Several approaches have been proposed, among them average log-likelihood \cite{theis2015generative}, Parzen window estimates \cite{breuleux2010unlearning} or visual fidelity of samples \cite{goodfellow2014generative} when suitable. In \cite{theis2015note} the authors show that these three criteria are largely independent of each other when the data is high-dimensional. In particular, they state that average likelihood is not a good measure.

In the field of image generating GANs, some more recently defined measures are the Inception Score (IS) \cite{salimans2016improved} and the Fréchet Inception Distance (FID) \cite{heusel2017gans}. Both approaches measure the distance between the original and the generated images. The Inception Score is computed as $\textrm{exp}(\mathbb{E}_x\textrm{\bf{KL}}(p(y|x) \parallel p(y)))$, where the Inception model \cite{szegedy2015going} is applied to every image to get the conditional label distribution $p(y|x)$. Images that contain meaningful objects are expected to have a conditional label distribution $p(y|x)$ with low entropy. On the other hand, it is expected that the images generated by the model have a degree of variation among them, so the marginal $\int p(y|x = G(z))dz$ should have high entropy. The Inception score is obtained from the combination of these two requirements, where the results are exponentiated so the values are easier to compare. \textbf{KL} stands for Kullback-Leibler divergence \cite{kullback1997information}. The Fréchet Inception Distance is computed as $d^2((M_r, \Sigma_r),(M_g, \Sigma_g)) = ||M_r – M_g||^2 + Tr(\Sigma_r + \Sigma_g - 2(\Sigma_r\Sigma_g)^\frac{1}{2})$, where $(M_r, \Sigma_r)$ and $(M_g, \Sigma_g)$ are the mean vectors and covariance matrices of the feature vectors for real and generated images, respectively. The feature vectors are computed as the values of the activation layer of the Inception model.

In layperson's terms, given two sets of images $I_A$ and $I_B$, the FID measures the similarity of the predictions of the Inception model over $I_a$ and $I_b$. FID is widely used as a performance measure in the image generation community, as in \cite{park2019semantic}\cite{wu2019logan}\cite{zhang2018stackgan++}\cite{nazeri2019edgeconnect}.

In \cite{borji2019pros} the author analyzes the pros and cons of several GAN performance measures. His work is focused on GAN applied to images, and arrives to the conclusion that FID score looks more plausible than others. Although it has its drawbacks, as to rely on pre-trained networks, which could pose problems when translated to other domains. However, as it is pointed out in a recent article \cite{barratt2018note}, ``there are no universally agreed-upon performance metrics for unsupervised learning, and people have already pointed out many shortcomings of these Inception-based methods. Until something better comes along though, they’re going to show up in every paper so it’s worth knowing what they are.'' 

Taken into account the relative quality of the FID score when applied to the image domain, one of the goals of this research is to find a way of adapting that score, based in a pre-trained model over a set of images, to sets of gestures.

\subsection{Applying Fréchet Gesture Distance to the baseline}
\label{subsec:fgd-to-baseline}
When trying to adapt the FID for gestures, the first problem is that, to the best of our knowledge, there is no model that could play the role of the Inception model. 
Let us remember that the Inception model has been created by a supervised deep learning algorithm and, when presented an input image, it outputs a set of probabilities of that image belonging to any of a thousand possible classes. For gestures, there is no such model. A possible approach could be to manually label a set of gestures, apply a supervised model over it, and then use it with the same role than the Inception model. As in our domain there is not a clear-cut classification of the gestures generated by the robot, we have chosen another alternative: to build a Gaussian Mixture Model (GMM) in an unsupervised fashion from a set of synthetic gestures created by Choregraphe\footnote{http://doc.aldebaran.com/2-5/software/choregraphe/index.html}, a software designed to create robot animations. It includes different type of predefined animations, such us body talk gestures, reactions and emotions, that are used to bring the robot to life. In a previous work \cite{rodriguez2019spontaneous}, we chose a set of animations from original Choregraphe's animation library that could be used as beat gestures, and we created a database with those animations. After sampling selected animations with a frequency of 4 Hz we obtained a database built up from 1502 poses. From now on we will refer to this database as Choregraphe gestures database (ChDB). In this approach, Choregraphe gestures play a similar role as the data from which the Inception model was created. As in the image domain a model independent from the analyzed data was created (Inception), in the gesture domain we create a model independent from the analyzed data (Choregraphe-based GMM). The data used by the GAN for training is different from Choregraphe data, as it is captured by a MoCap system.

The GMM election is supported by previous motion work by the authors \cite{rodriguez2019spontaneous}, where they show that this model ranks second after GANs in the quality of generated gestures, when used as generative models. When evaluating the quality of the gestures created by the GAN, the computed GMM model can be used to classify new gestures, and thus return the set of probabilities needed to compute the FID.

The process to define the Fréchet Gesture Distance (FGD) between two sets of gestures $G_A$ and $G_B$ is the following:

\begin{itemize}
    \item Create a database $G_M$ from Choregraphe gestures.
    \item Build a GMM from $G_M$.
    \item Compute the probabilities $P(G_A)$ and $P(G_B)$ returned by that GMM over $G_A$ and $G_B$, respectively.
    \item {Compute the Fréchet distance over $P(G_A)$ and $P(G_B)$.}
\end{itemize}

The GMM has been built with 24 components sharing the same covariance matrix. After a initial Choregraphe gestures database (ChDB) of $K = 1502$ poses is built, it needs to be adapted for the different sizes of the unit of movement. When $\mu = 1$ the length of the pose and the unit of movement are the same, and the Choregraphe gestures database can be used without further processing. Therefore, if we denote as ChDB-n the Choregraphe database used when $\mu = n$, we find that ChDB-1 and ChDB are the same. But, in the general case, with an arbitrary value of $\mu = n$, the dimensionality of ChDB-n is $14 \times n$. To achieve this, $n$ consecutive poses are joined together in ChDB-n, thus bringing the number of units of movement in ChDB-n to $K / n$. Therefore, the GMM is trained with the Chdb-$\mu$ associated to each value of the $\mu$ parameter.

Table~\ref{tab:fid-res} shows the FGD distance values for the different $\mu$ values. According to these values $\mu=4$ shows the shortest distance and thus it seems the most adequate value.
\begin{table}[!htbp]
\caption{FGD values for the different comparisons}
\label{tab:fid-res}
\begin{tabular}[!htbp]{lc|cc|}\hline
UM  & & OpenNI GAN & OpenPose GAN\\ \hline
 \multirow{2}{*}{$\mu$ = 4} &  $E$ & 0.1309  & 0.1231 \\
   &  $\sigma$ & 0.0117
 &  0.0108 \\  \hline
\multirow{2}{*}{$\mu$ = 6} &  $E$ &  0.2452 & 0.1773\\
   &  $\sigma$ & 0.0154 & 0.0131 \\  \hline
\multirow{2}{*}{$\mu$ = 8} &  $E$ & 0.5425  & 0.2399\\
   &  $\sigma$ & 0.0209 & 0.01457 \\   \hline
\vspace{+1mm}
\end{tabular}
\end{table}

\section{Visual evaluation of the behavior}
\label{sec:visual-evaluation}
\begin{figure}[!htbp]
   \subfigure[OpenNI\label{fig:openni-replica}]{
     \includegraphics[width=1.0\columnwidth]{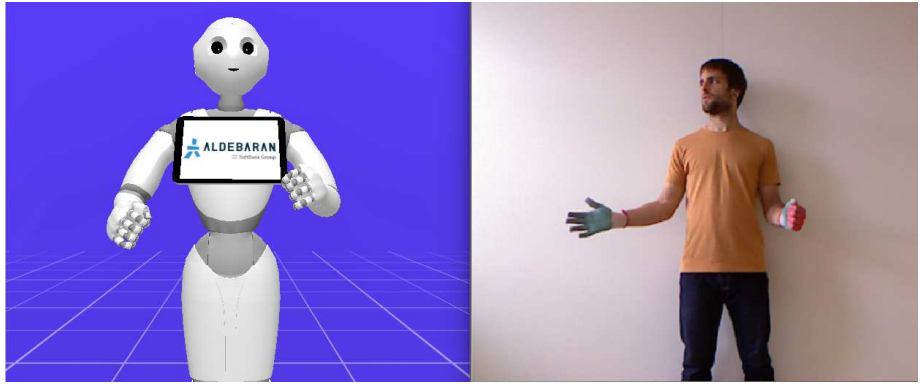}}
   \subfigure[OpenPose\label{fig:openpose-replica}]{
     \includegraphics[width=1.0\columnwidth]{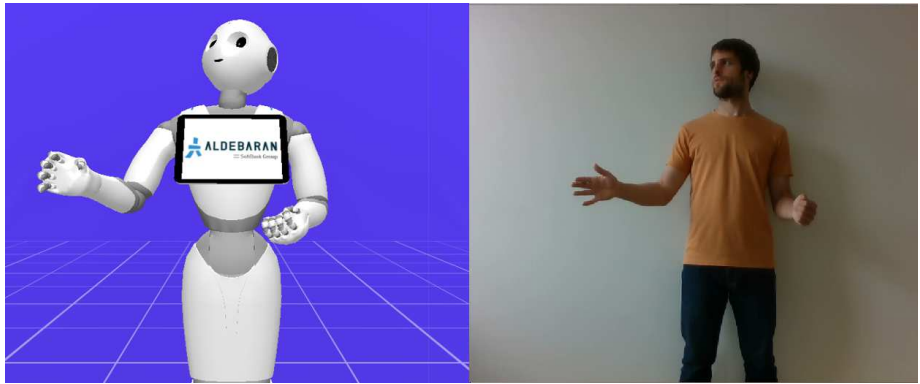}}
\caption{Reproduction of poses in the simulated robot}
\label{fig:replicas}
\end{figure}

Visual inspection of the robot behavior can be somewhat subjective, specially when variations are subtle. However, the robot behavior must be perceived as acceptable by humans in any circumstance. The two approaches compared in this work are very similar in nature, the only difference being the MoCap system used to generate the learning data. The main differences between them were the difficulties to accurately track the head and hands positions with OpenNI. Figure~\ref{fig:replicas} shows those differences. The reader is invited to pay attention to how the head and hand positions differ.

These difficulties are therefore reflected in the generated gestures, as can be appreciated in this video\footnote{https://www.youtube.com/watch?v=h9wpMEH8JQc}.
The executions of both systems correspond to the models trained to generate movements using $\mu$=4 as unit of movement. Notice that the temporal length of the audio intended to be pronounced by the robot determines the number of UM required to the generative model. Thus, the execution of those UMs, one after the other, defines the whole movement displayed by the robot.
On the one hand, head information provided by the OpenNI skeleton tracking package was not enough for preserving head movements and thus, the resulting motion was poor. On the other hand, the tracker only offered wrist positions and as a consequence, a vision based alternative was developed by segmenting red/green colors of the gloves wore by the speaker for tracking palms and backs of both hands. The opening/closing of the fingers was made at random for each generated movement. Lastly, the robot elbows tended to be too separated from the body and raised up. 
At a glance, it can be seen that the OpenPose based approach overcomes these three main drawbacks.

\section{Discussion and further work}
\label{sec:further-work}

In this paper an approach to quantitatively measure the degree of fidelity/ori\-gi\-na\-li\-ty of a gesture generating method is presented. Two beat gestures generation approaches are compared: OpenNI based GAN model and OpenPose based GAN model. These two systems basically differ in the MoCap system being used for acquiring the database used for learning the generative model.

To measure the fidelity of the generated samples to the original ones we performed a PCoA over the original and generated samples for the two types of MoCap and different length of units of movements. The visual analysis,  as well as the decomposition of the variances  in this step support the hypothesis that the generated gestures indeed are similar to the original ones. We also discovered  that the explained variances by the regression models to recover the original units are bigger in OpenPose, which could point to bigger fidelity to the original. In the same  vein UM 4 and 6 appear to have higher degree of fidelity. To measure the originality, we calculated procrustes statistics and we observed that in general terms, the originality seems bigger in OpenPose and at the same time this approach generates smoother movements according to two motion measures: jerk and length path.

Finally, we have defined a Fréchet Gesture Distance (FGD) which is inspired in the Fréchet Inception Distance (FID) to be able to see how far are the generated gestures from the original ones. 
The Fréchet distance is a measure of the similarity between two distributions and in our context those two distributions are the probabilities assigned by a classifier over all the possible classes when presented a new instance. Therefore, FGD is generator-agnostic, in the sense that it is irrelevant how the objects have been created, only their predicted probabilities when applying some model (as with Inception in the case of FID) are taken into account.

Let us remember that we want the generated gestures to be similar, but not too much. We could wonder if, given the data collected so far (PCoA, jerk), the similarity constraint has been already fulfilled (let us remember that the difference in variance composition tips the balance in the other direction), and the FGD will be bigger (more different). To pursue this analysis, we have computed FGD for the two MoCaps (OpenNI and OpenPose) and three types of number of units of movements (4, 6 and 8). We see that FGD for OpenPose is smaller than for OpenNI, so it seems reasonable to suppose that in the balance between similarity and originality, the smaller the FGD measure the better. This leads us also to the conclusion that 4 units of movements are better than 6 or 8.

Visual inspection reflected that although subtle the difficulties to track hands and head positions were translated to the GAN generated gestures. And subtle are also the differences among the measured values, probably because the two systems being compared are equal in nature. Thus, as further work we plan to repeat the analysis to observe if the results of these different quantitative techniques are translatable when comparing, for instance, GAN based approaches to other motion generation approaches such as variational autoencoders.

\bibliographystyle{spmpsci}
\bibliography{ms}

\begin{thebibliography}{10}
\providecommand{\url}[1]{{#1}}
\providecommand{\urlprefix}{URL }
\expandafter\ifx\csname urlstyle\endcsname\relax
  \providecommand{\doi}[1]{DOI~\discretionary{}{}{}#1}\else
  \providecommand{\doi}{DOI~\discretionary{}{}{}\begingroup
  \urlstyle{rm}\Url}\fi

\bibitem{alibeigi2017inverse}
Alibeigi, M., Rabiee, S., Ahmadabadi, M.N.: Inverse kinematics based human
  mimicking system using skeletal tracking technology.
\newblock Journal of Intelligent {\&} Robotic Systems \textbf{85}(1), 27--45
  (2017)

\bibitem{barratt2018note}
Barratt, S., Sharma, R.: A note on the inception score.
\newblock arXiv preprint arXiv:1801.01973  (2018)

\bibitem{beck2017body}
Beck, A., Yumak, Z., Magnenat-Thalmann, N.: Body movements generation for
  virtual characters and social robots.
\newblock In: Social signal processing, chap.~20, pp. 273--286. Cambridge
  University Press (2017)

\bibitem{becker2011evaluating}
{Becker-Asano}, C., {Ishiguro}, H.: Evaluating facial displays of emotion for
  the android robot {Geminoid F}.
\newblock In: 2011 IEEE Workshop on Affective Computational Intelligence
  (WACI), pp. 1--8 (2011).
\newblock \doi{10.1109/WACI.2011.5953147}

\bibitem{borji2019pros}
Borji, A.: Pros and cons of {GAN} evaluation measures.
\newblock Computer Vision and Image Understanding \textbf{179}, 41--65 (2019)

\bibitem{breuleux2010unlearning}
Breuleux, O., Bengio, Y., Vincent, P.: Unlearning for better mixing.
\newblock Universite de Montreal/DIRO  (2010)

\bibitem{calinon04learning}
Calinon, S., D'halluin, F., Sauser, E.L., Cakdwell, D.G., Billard, A.G.:
  Learning and reproduction of gestures by imitation.
\newblock In: International Conference on Intelligent Robots and Systems, pp.
  2769--2774 (2004)

\bibitem{cao2018openpose}
Cao, Z., Hidalgo, G., Simon, T., Wei, S.E., Sheikh, Y.: Open{P}ose: realtime
  multi-person 2{D} pose estimation using {P}art {A}ffinity {F}ields.
\newblock In: arXiv preprint arXiv:1812.08008 (2018)

\bibitem{carpinella2017rosas}
Carpinella, C., Wyman, A., Perez, M., Stroessner, S.: The robotic social
  attributes scale ({RoSAS}): Development and validation.
\newblock In: 17th Human Robot Interaction, pp. 254--262 (2017).
\newblock \doi{10.1145/2909824.3020208}

\bibitem{cerrato2017engagement}
Cerrato, L., Campbell, N.: Engagement in dialogue with social robots.
\newblock In: Dialogues with Social Robots, pp. 313--319. Springer (2017)

\bibitem{eielts2020cothought}
Eielts, C., Pouw, W., Ouwehand, K., van Gog, T., Zwaan, R.A., Paas, F.:
  Co-thought gesturing supports more complex problem solving in subjects with
  lower visual working-memory capacity.
\newblock Psychological Research \textbf{84}(2), 502--513 (2020).
\newblock \doi{10.1007/s00426-018-1065-9}

\bibitem{gao2016calibration}
Gao, X., Yun, C., Jin, H., Gao, Y.: Calibration method of robot base frame
  using procrustes analysis.
\newblock In: 2016 Asia-Pacific Conference on Intelligent Robot Systems
  (ACIRS), pp. 16--20. IEEE (2016)

\bibitem{goodfellow2014generative}
Goodfellow, I., Pouget-Abadie, J., Mirza, M., Xu, B., Warde-Farley, D., Ozair,
  S., Courville, A., Bengio, Y.: Generative adversarial nets.
\newblock In: Advances in neural information processing systems, pp. 2672--2680
  (2014)

\bibitem{gower85encyclopedia}
Gower, J.: Encyclopedia of Statistical Sciences, vol.~5, chap. Measures of
  similarity, dissimilarity and distance, pp. 397--405.
\newblock John Wiley \& Sons, New York (1985)

\bibitem{gower1966some}
Gower, J.C.: Some distance properties of latent root and vector methods used in
  multivariate analysis.
\newblock Biometrika \textbf{53}(3--4), 325--338 (1966)

\bibitem{gower2004procrustes}
Gower, J.C., Dijksterhuis, G.B., et~al.: Procrustes problems, vol.~30.
\newblock Oxford University Press on Demand (2004)

\bibitem{hasegawa2018evaluation}
Hasegawa, D., Kaneko, N., Shirakawa, S., Sakuta, H., Sumi, K.: Evaluation of
  speech-to-gesture generation using bi-directional {LSTM} network.
\newblock In: 18th International Conference on Intelligent Virtual Agents, pp.
  79--86 (2018).
\newblock \doi{10.1145/3267851.3267878}

\bibitem{heusel2017gans}
Heusel, M., Ramsauer, H., Unterthiner, T., Nessler, B., Hochreiter, S.: {GANs
  trained by a two time-scale update rule converge to a local Nash
  equilibrium}.
\newblock In: Advances in neural information processing systems, pp. 6626--6637
  (2017)

\bibitem{hotelling1993analysis}
Hotelling, H.: Analysis of a complex of statistical variables into principal
  components.
\newblock Journal of Educational Psychology \textbf{24}(6), 417–441 (1993).
\newblock \doi{https://doi.org/10.1037/h0071325}

\bibitem{jarque19}
Jarque-Bou, N.J., Scano, A., Atzori, M., M{\"u}ller, H.: Kinematic synergies of
  hand grasps: a comprehensive study on a large publicly available dataset.
\newblock Journal of neuroengineering and rehabilitation \textbf{16}(1), 63
  (2019)

\bibitem{kofinas2015complete}
Kofinas, N., Orfanoudakis, E., Lagoudakis, M.G.: Complete analytical forward
  and inverse kinematics for the nao humanoid robot.
\newblock Journal of Intelligent {\&} Robotic Systems \textbf{77}(2), 251--264
  (2015).
\newblock \doi{10.1007/s10846-013-0015-4}.
\newblock \urlprefix\url{https://doi.org/10.1007/s10846-013-0015-4}

\bibitem{kucherenko2019importance}
Kucherenko, T., Hasegawa, D., Kaneko, N., Henter, G., Kjellström, H.: On the
  importance of representations for speech-driven gesture generation.
\newblock In: 18th International Conference on Autonomous Agents and MultiAgent
  Systems (AAMAS), pp. 2072--2074 (2019)

\bibitem{kucherenko2020gesticulator}
{Kucherenko}, T., {Jonell}, P., {van Waveren}, S., {Eje Henter}, G.,
  {Alexanderson}, S., {Leite}, I., {Kjellstr{\"o}m}, H.: {Gesticulator: A
  framework for semantically-aware speech-driven gesture generation}.
\newblock arXiv e-prints arXiv:2001.09326 (2020)

\bibitem{kullback1997information}
Kullback, S.: Information theory and statistics.
\newblock Courier Corporation (1997)

\bibitem{lhommet15expressing}
Lhommet, M., Marsella, S.: The Oxford Handbook of Affective Computing, chap.
  Expressing Emotion Through Posture and Gesture, pp. 273--285.
\newblock Oxford University Press (2015)

\bibitem{makondo2015knowledge}
Makondo, N., Rosman, B., Hasegawa, O.: Knowledge transfer for learning robot
  models via local procrustes analysis.
\newblock In: 2015 IEEE-RAS 15th International Conference on Humanoid Robots
  (Humanoids), pp. 1075--1082. IEEE (2015)

\bibitem{maset2020procrustes}
Maset, E., Scalera, L., Zonta, D., Alba, I., Crosilla, F., Fusiello, A.:
  Procrustes analysis for the virtual trial assembly of large-size elements.
\newblock Robotics and Computer-Integrated Manufacturing \textbf{62}, 101885
  (2020)

\bibitem{mcneill1992hand}
McNeill, D.: Hand and mind: What gestures reveal about thought.
\newblock University of Chicago press (1992)

\bibitem{mukherjee2015inverse}
Mukherjee, S., Paramkusam, D., Dwivedy, S.K.: Inverse kinematics of a {NAO}
  humanoid robot using {Kinect} to track and imitate human motion.
\newblock In: International Conference on Robotics, Automation, Control and
  Embedded Systems {(RACE)}. {IEEE} (2015)

\bibitem{nazeri2019edgeconnect}
Nazeri, K., Ng, E., Joseph, T., Qureshi, F.Z., Ebrahimi, M.: Edgeconnect:
  Generative image inpainting with adversarial edge learning.
\newblock arXiv preprint arXiv:1901.00212  (2019)

\bibitem{pan2018evaluating}
Pan, M., Croft, E., Niemeyer, G.: Evaluating social perception of
  human-to-robot handovers using the robot social attributes scale (rosas).
\newblock In: ACM/IEEE International Conference on Human-Robot Interaction
  (HRI), p. 443–451 (2018).
\newblock \doi{https://doi.org/10.1145/3171221.3171257}

\bibitem{park15}
Park, G., Konno, A.: Imitation learning framework based on principal component
  analysis.
\newblock Advanced Robotics \textbf{29}(9), 639--656 (2015).
\newblock \doi{10.1080/01691864.2015.1007084}.
\newblock \urlprefix\url{https://doi.org/10.1080/01691864.2015.1007084}

\bibitem{park2019semantic}
Park, T., Liu, M.Y., Wang, T.C., Zhu, J.Y.: Semantic image synthesis with
  spatially-adaptive normalization.
\newblock In: Proceedings of the IEEE Conference on Computer Vision and Pattern
  Recognition, pp. 2337--2346 (2019)

\bibitem{penna2013whole}
Poubel, L.P.: Whole-body online human motion imitation by a humanoid robot
  using task specification.
\newblock Master's thesis, Ecole Centrale de Nantes--Warsaw University of
  Technology (2013)

\bibitem{rodriguez2014humanizing}
Rodriguez, I., Astigarraga, A., Jauregi, E., Ruiz, T., Lazkano, E.: Humanizing
  {NAO} robot teleoperation using {ROS}.
\newblock In: International Conference on Humanoid Robots (Humanoids) (2014)

\bibitem{rodriguez2019spontaneous}
Rodriguez, I., Mart\'{\i}nez-Otzeta, J.M., Irigoien, I., Lazkano, E.:
  Spontaneous talking gestures using generative adversarial networks.
\newblock Robotics and Autonomous Systems \textbf{114}, 57 -- 65 (2019)

\bibitem{salimans2016improved}
Salimans, T., Goodfellow, I., Zaremba, W., Cheung, V., Radford, A., Chen, X.:
  Improved techniques for training {GANs}.
\newblock In: Advances in neural information processing systems, pp. 2234--2242
  (2016)

\bibitem{suguitan2020moveae}
Suguitan, M., Gomez, R., Hoffman, G.: Move{AE}: Moditying affective robot
  movements using classifying variational autoencoders.
\newblock In: {ACM/IEEE} International Conference on Human Robot Interaction
  (HRI), pp. 481--489 (2020).
\newblock \doi{10.1145/3267851.3267878}

\bibitem{szegedy2015going}
Szegedy, C., Liu, W., Jia, Y., Sermanet, P., Reed, S., Anguelov, D., Erhan, D.,
  Vanhoucke, V., Rabinovich, A.: Going deeper with convolutions.
\newblock In: Proceedings of the IEEE conference on computer vision and pattern
  recognition, pp. 1--9 (2015)

\bibitem{theis2015generative}
Theis, L., Bethge, M.: Generative image modeling using spatial lstms.
\newblock In: Advances in Neural Information Processing Systems, pp. 1927--1935
  (2015)

\bibitem{theis2015note}
Theis, L., van~den Oord, A., Bethge, M.: A note on the evaluation of generative
  models.
\newblock CoRR \textbf{abs/1511.01844} (2015)

\bibitem{velner2020intonation}
Velner, E., Boersma, P.P., de~Graaf, M.M.: Intonation in robot speech: Does it
  work the same as with people?
\newblock In: {ACM/IEEE} International Conference on Human-Robot Interaction
  (HRI), pp. 569--578 (2020)

\bibitem{wolfert2019should}
Wolfert, P., Kucherenko, T., Kjelström, H., Belpaeme, T.: Should beat gestures
  be learned or designed? a benchmarking user study.
\newblock In: ICDL-EPIROB 2019 Workshop on Naturalistic Non-Verbal and
  Affective Human-Robot Interactions, p.~4 (2019)

\bibitem{wood18}
Wood, M., Simmatis, L., Boyd, J.G., Scott, S., Jacobson, J.: Using principal
  component analysis to reduce complex datasets produced by robotic technology
  in healthy participants.
\newblock Journal of NeuroEngineering and Rehabilitation \textbf{15} (2018).
\newblock \doi{10.1186/s12984-018-0416-5}

\bibitem{wu2019logan}
Wu, Y., Donahue, J., Balduzzi, D., Simonyan, K., Lillicrap, T.: {LOGAN}: Latent
  optimisation for generative adversarial networks.
\newblock arXiv preprint arXiv:1912.00953  (2019)

\bibitem{zabala2019learning}
Zabala, U., Rodriguez, I., Mart\'{\i}nez-Otzeta, J.M., Lazkano, E.: Learning to
  gesticulate by observation using a deep generative approach.
\newblock In: 11th International Conference on Social Robotics (ICSR) (2019
  (Accepted)).
\newblock \urlprefix\url{http://arxiv.org/abs/1909.01768}

\bibitem{zhang2018stackgan++}
Zhang, H., Xu, T., Li, H., Zhang, S., Wang, X., Huang, X., Metaxas, D.N.:
  Stackgan++: Realistic image synthesis with stacked generative adversarial
  networks.
\newblock IEEE transactions on pattern analysis and machine intelligence
  \textbf{41}(8), 1947--1962 (2018)

\bibitem{zhang2019real}
Zhang, Z., Niu, Y., Kong, L.D., Lin, S., Wang, H.: A real-time upper-body robot
  imitation system.
\newblock International Journal of Robotics and Control \textbf{2}, 49--56
  (2019).
\newblock \doi{10.5430/ijrc.v2n1p49}

\end{thebibliography}
\end{document}